%% file: acl_latex.tex
\pdfoutput=1
\documentclass[11pt]{article}

\usepackage[preprint]{acl}

\usepackage{times}
\usepackage{latexsym}
\usepackage{algpseudocode}
\usepackage{algorithm}
\usepackage[T1]{fontenc}
\usepackage[utf8]{inputenc}

\usepackage{microtype}

\usepackage{inconsolata}

\usepackage{graphicx}
\usepackage[acronym]{glossaries}

\usepackage{inconsolata}

\usepackage{multicol}
\usepackage{booktabs}
\usepackage{makecell}
\usepackage{amsmath, amssymb}
\usepackage{multirow}
\usepackage{mathtools}
\usepackage{xcolor}
\usepackage{enumitem}
\usepackage{float}
\usepackage{graphicx}
\usepackage{upgreek}
\usepackage{seqsplit}
\usepackage{color,soul}
\usepackage{arydshln}
\usepackage{placeins}
\usepackage{pifont}
\usepackage{amssymb}
\usepackage{bbm}

\input{math_commands}

\usepackage{caption}
\usepackage{subcaption}
\title{AT-RAG: An Adaptive RAG Model Enhancing Query Efficiency with Topic Filtering and Iterative Reasoning}

\author{Mohammad R. Rezaei, Maziar Hafezi,\\ {\bf Amit Satpathy, Lovell Hodge, Ebrahim Pourjafari\textsuperscript{*}} \\
        Munich Reinsurance Co-Canada, Toronto, Ontario, Canada  \\  \textsuperscript{*}\texttt{epourjafari@munichre.ca}}

\begin{document}
\newacronym{mm}{AT-RAG}{Adaptive Topic RAG}
\newacronym{llm}{LLM}{Large Language Model}
\newacronym{cot}{CoT}{Chain-of-Thought}
\newacronym{qa}{QA}{Question Answering}
\newacronym{rag}{RAG}{Retrieval Augmented Generation }
\maketitle
\begin{abstract}
Recent advancements in \gls{qa} with \glspl{llm} like GPT-4 have shown limitations in handling complex multi-hop queries. We propose \gls{mm}, a novel multistep \gls{rag}, which incorporates topic modeling for efficient document retrieval and reasoning. Using BERTopic, our model dynamically assigns topics to queries, improving retrieval accuracy and efficiency.
We evaluated \gls{mm} on multihop benchmark datasets (\gls{qa}) and a medical case study \gls{qa}. Results show significant improvements in correctness, completeness, and relevance compared to existing methods. \gls{mm} reduces retrieval time while maintaining high precision, making it suitable for general tasks \gls{qa} and complex domain-specific challenges such as medical \gls{qa}.
The integration of topic filtering and iterative reasoning enables our model to handle intricate queries efficiently, which makes it suitable for applications that require nuanced information retrieval and decision-making.

\end{abstract}

\input{sec-introduction}
\input{sec-related}
\input{sec-method}
\input{sec-experiments}
\input{sec-application}
\input{sec-discussion}

\section*{Acknowledgements}
% \section*{References}

\bibliography{main}
\newpage
\appendix
\input{sec-appendix}

\end{document}

%% file: math_commands.tex
\usepackage{amsmath,amsfonts,bm}

\def\eqref#1{equation~\ref{#1}}
\def\1{\bm{1}}

\def\rva{{\mathbf{a}}}

\def\rvc{{\mathbf{c}}}
\def\rvd{{\mathbf{d}}}

\def\rvq{{\mathbf{q}}}

\def\rvx{{\mathbf{x}}}
\def\rvy{{\mathbf{y}}}

\DeclareMathAlphabet{\mathsfit}{\encodingdefault}{\sfdefault}{m}{sl}
\SetMathAlphabet{\mathsfit}{bold}{\encodingdefault}{\sfdefault}{bx}{n}

%% file: sec-introduction.tex
\section{Introduction}

\gls{llm}s have transformed natural language processing, particularly in \gls{qa} tasks, by generating coherent and contextually relevant responses using their vast pre-trained knowledge \citep{achiam2023gpt,team2023gemini,jiang2024mixtral}. Although models like GPT-4 demonstrate impressive capabilities, they face significant challenges in responding to queries that require external information or reasoning across multiple documents\citep{raiaan2024review,kwiatkowski2019natural}. These limitations are especially evident in multi-hop \gls{qa} scenarios, where extracting and synthesizing information from various sources is essential for producing accurate answers\citep{press2022measuring,tang2024multihop}. To address these challenges, \gls{rag} models have been developed to enable \gls{llm}s to access relevant external knowledge and enhance the quality of responses.

\begin{figure}[tb]
\includegraphics[width=1\columnwidth]{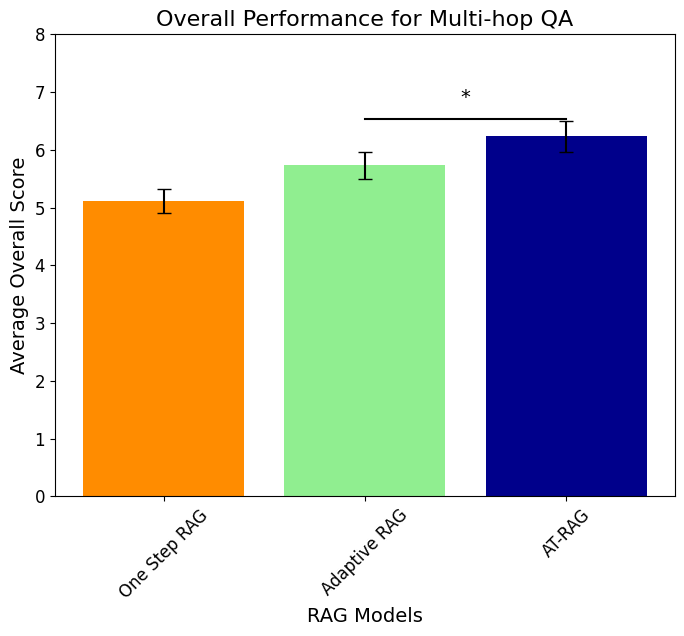}
    \caption{Comparison of the average overall score across multiple datasets for different \gls{rag} models (One Step \gls{rag}, Adaptive \gls{rag}, \gls{mm} with GPT40). Error bars depict the standard deviations for each model. An ANOVA test \citep{st1989analysis} (with p<0.05) reveals a statistically significant difference between the \gls{mm} and Adaptive \gls{rag}, denoted by an asterisk (*). For further details, refer to Table \ref{tab:main}}
    \label{fig:performance}
\end{figure}

In this paper, we introduce \gls{mm}, a novel multi-step retrieval-based \gls{qa} framework that enhances the retrieval process by incorporating a topic assignment model. This model filters external knowledge in \gls{qa} tasks by assigning relevant topics to each query, ensuring retrieval focuses on contextually significant information. This approach improves retrieval accuracy and reduces the computational overhead associated with multi-step retrieval processes. Furthermore, \gls{mm} integrates \gls{cot} reasoning \citep{wei2022chain,wang2024chainofthought}, allowing iterative document retrieval and reasoning to handle complex multi-hop queries better.

We evaluated \gls{mm} on several challenging multi-hop \gls{qa} datasets, including \textbf{HotpotQA} \citep{yang2018enhancing}, \textbf{MuSiQue} \citep{trivedi2022interleaving}, and \textbf{2WikiMultiHopQA} \citep{ho2020constructing}. Our results show that \gls{mm} outperforms existing \gls{rag} models in terms of accuracy; see Figure \ref{fig:performance}. This work's main contribution is the introduction of topic-guided retrieval aimed at enhancing multi-step reasoning. The model undergoes evaluation on intricate datasets, with a thorough analysis of improvements in answer quality.

The experimental results reveal that \gls{mm} significantly improves accuracy compared to existing methods like Adaptive-\gls{rag} \citep{lewis2020retrieval}. These improvements are particularly pronounced in multi-hop reasoning tasks, where prior adaptive strategies often incur high computational costs. By narrowing the search space through a stochastic topic assignment process, \gls{mm} reduces the number of documents needed for retrieval while improving the accuracy of query resolution. We evaluated \gls{mm} using various state-of-the-art \glspl{llm}, including GPT-4 and Mixtral8x7B \citep{jiang2024mixtral}. \gls{mm} demonstrated superior performance when leveraging GPT-4 as the \gls{llm}, proving particularly effective in addressing complex multi-hop queries with optimized retrieval and precise reasoning.

Furthermore, we conducted a case study to assess the effectiveness of \gls{mm} in addressing multi-hop queries in the real world, focusing on answering medical questions. Medical records, which are longitudinal and comprise various documents such as doctor's notes, lab results, diagnostics, and medications, present a unique challenge for the \gls{rag} frameworks. Time-based queries, such as \textit{ 'What are the abnormal laboratory results of the patient in the last year?'}, often yield suboptimal results when processed by a naive \gls{rag}. These approaches struggle to retrieve relevant information from potentially thousands of document chunks within a vector database. Furthermore, naive \gls{rag} may lack the reasoning ability to identify the correct time-stamped data that meet both the time range and the specific condition. As illustrated in Fig. \ref{fig:medical}, our method consistently outperforms naive \gls{rag} approaches across all six evaluated cases.

%% file: sec-related.tex
\section{Related Work}

Recent advancements in multi-hop \gls{qa} systems have emphasized enhancing accuracy, efficiency, and reasoning capabilities. This section reviews three main categories: question decomposition, \gls{cot} with iterative retrieval, and adaptive retrieval.

\subsection{Question Decomposition}
Question decomposition breaks down complex queries into simpler sub-questions for a more straightforward resolution. Khattab et al. \citep{shao2023enhancing} proposed the Iterative Retriever, Reader, and Reranker framework, which decomposes queries, retrieves relevant passages, and synthesizes information to generate answers \citep{zhang2024accelerating}. Press et al. (2023) introduced "Decomposed Prompting," a technique that leverages large language models to simplify complex queries into more manageable sub-questions\citep{schulhoff2024prompt}.

\subsection{Chain-of-Thought with Iterative Retrieval}
This approach combines \gls{cot} reasoning with iterative document retrieval. Yao et al. \citep{sun2022recitation} introduced "ReCite," where a large language model generates reasoning steps while retrieving relevant documents iteratively. Generating intermediate reasoning steps improves the model's ability to handle complex reasoning tasks. Furthermore, such reasoning capabilities naturally emerge in large models through \gls{cot} prompting \citep{wei2022chain}.

\subsection{Adaptive Retrieval}
Adaptive retrieval methods dynamically adjust the retrieval process based on the specific needs of each query \citep{fan2024survey}. \citep{asai2023self} introduced a system that allows a language model to iteratively formulate and resolve subsequent queries as needed. Another notable approach is IRCoT, which integrates the retrieval and reasoning phases, improving multi-step question answering on datasets like HotpotQA and 2WikiMultiHopQA \citep{trivedi2022interleaving}. The Adaptive-RAG framework takes this further by selecting the most appropriate retrieval strategy based on the complexity of the query.

Despite these advancements, challenges remain. Current systems lack a fully flexible approach that dynamically adapts retrieval and reasoning processes in response to query complexity. Balancing efficiency and thoroughness, especially for queries with varying difficulty levels, continues to be a critical area for improvement. Future work must address these issues to enhance the adaptability and performance of multi-hop \gls{qa} systems.

%% file: sec-method.tex
\section{Method}

This section introduces the proposed \gls{rag} model, \gls{mm}, which combines single-step and multi-step retrieval strategies enhanced by topic assignment to tackle complex \gls{qa} challenges. By integrating topic modeling with multi-step reasoning, the model improves both retrieval precision and efficiency, leading to more accurate and well-reasoned answers.

\subsection{Background}
\glspl{llm} is designed to process an input sequence of tokens and generate an output sequence. Formally, given an input sequence $\rvx = [x_1, x_2, \dots, x_n]^T$, the \gls{llm} generates an output sequence $\rvy = [y_1, y_2, \dots, y_m]^T$, expressed as $\rvy = LLM(\rvx)$ where $n$ and $m$ are the number of tokens for each sequence. In the context of \gls{qa}, the input sequence $\rvx$ corresponds to the user's query $\rvq$, and the output sequence $\rvy$ corresponds to the generated answer $\hat{\rva}$, defined as $\hat{\rva} = LLM(\rvq)$. Ideally, $\hat{\rva}$ should match the correct answer $\rva$. 

While this non-retrieval-based \gls{qa} method is efficient and leverages the vast knowledge within the \gls{llm}, it struggles with queries requiring precise or up-to-date information, such as details about specific people or events beyond the \gls{llm}’s internal knowledge. Non-retrieval \gls{qa} is effective for simple queries but faces limitations with more complex or niche questions.

\subsubsection{One Step RAG for QA}

To overcome the limitations of nonretrieval methods for queries requiring external knowledge, retrieval-based methods \gls{qa} can be employed. This method utilizes external knowledge $\rvd$, retrieved from a knowledge source $\mathbf{D}$ (e.g., Wikipedia \citep{chen2017reading} or Wikidata \citep{vrandevcic2014wikidata}), which contains millions of documents. The retrieval process is formalized as:
\[
\rvd = Retriever(\rvq; \mathbf{D})
\]
where \texttt{Retriever} is the model that searches $\mathbf{D}$ for relevant documents based on query $\rvq$. The retrieved knowledge $\rvd$ is then incorporated into the input of \gls{llm}, enhancing the \gls{qa} process by generating an answer $\hat{\rva}$ based on both the query and the retrieved documents:
\[
\hat{\rva} = LLM(\rvq, \rvd)
\]
This approach improves the performance of the \gls{llm} for queries requiring specific or real-time information, augmenting its pre-trained knowledge with external sources.

\subsubsection{Multi-Step RAG for QA}

Although single-step \gls{rag} is effective for many queries, it has limitations when dealing with complex questions requiring simultaneous processing of information in multiple documents or the reasoning of interconnected knowledge. A, a multi-step \gls{rag} approach is introduced to address complex question, where the \gls{llm} iteratively interact with the retriever.

At each iteration $i=1,...,N$, the retriever fetches new documents $\rvd_i$ from $\mathbf{D}$, and the \gls{llm} incorporates both the newly retrieved document $\rvd_i$ and the context $\rvc_i$ (which includes previously retrieved documents and intermediate answers $\hat{\rva}_1, \hat{\rva}_2, \dots, \hat{\rva}_{i-1}$):
\begin{equation}
\hat{\rva}_i = LLM(\rvq, \rvd_i, \rvc_i)
\end{equation}
\begin{equation*}
\rvd_i = Retriever(\rvq, \rvc_i; \mathbf{D})
\end{equation*}
This iterative process continues until the \gls{llm} constructs a comprehensive understanding of the query, leading to the final answer. This approach is beneficial for complex multi-hop queries, where information needs to be integrated from multiple retrievals. However, it is more resource-intensive due to the increased computational cost of repeated interactions between the retriever and the \gls{llm}.

\subsection{Topic Filtering for RAG }
\begin{figure}[ht]
\includegraphics[width=0.8\columnwidth]{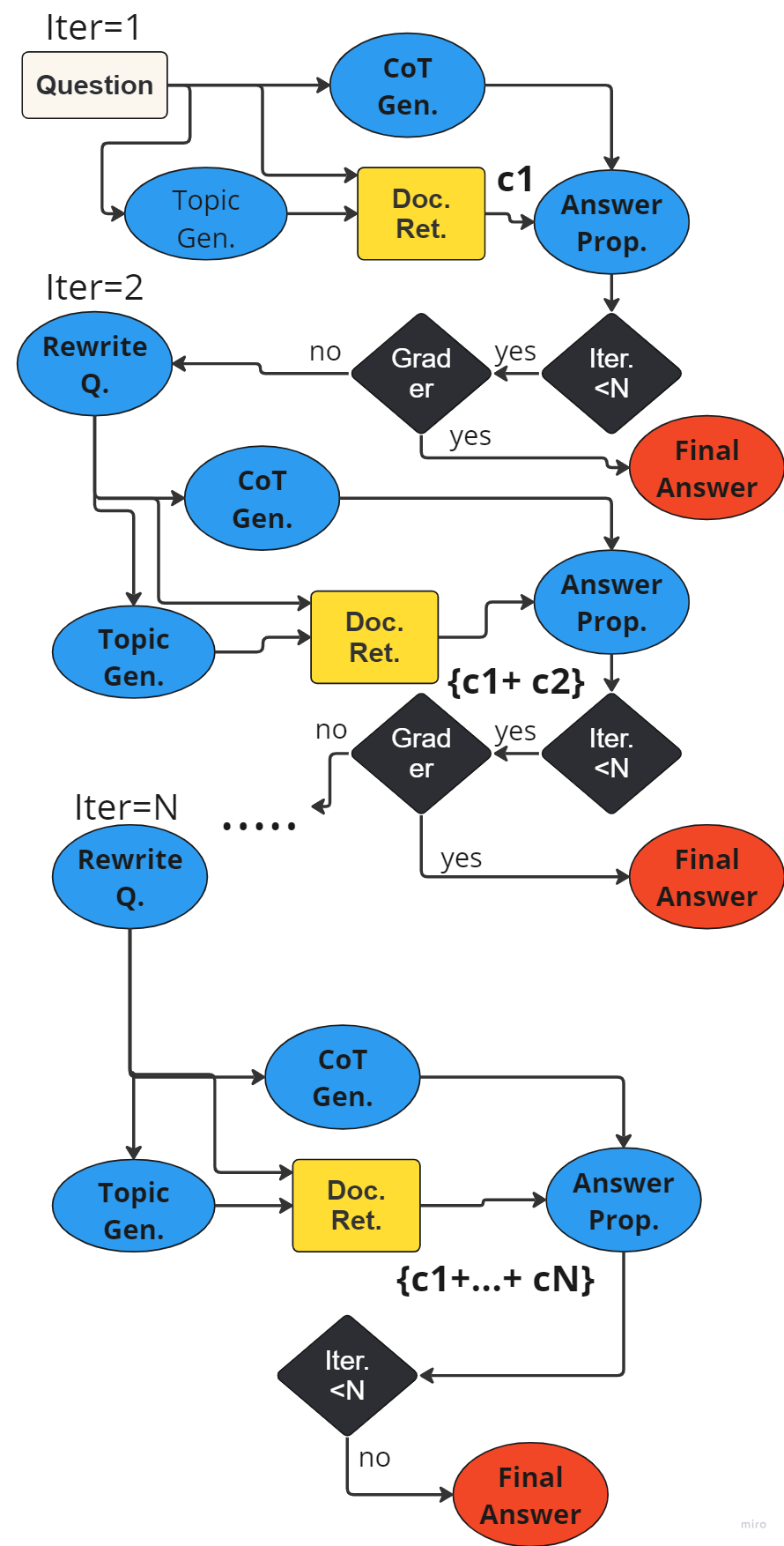}
    \caption{The \gls{mm} answering pipeline leverages a topic generator to streamline document retrieval. It iteratively generates reasoning steps through a \gls{cot} generator, guiding the formulation of answers. This process alternates between retrieval and reasoning until a predefined maximum number of iterations (N) is reached or the answer passes quality checks by grader nodes.}
    \label{fig:model}
\end{figure}
To improve the efficiency of Multi-Step \gls{rag} in \gls{qa}, we propose the \gls{mm} model. Before passing the query to the retriever, it is processed by a topic assignment module that generates a topic based on the input query. This topic reduces the search space by filtering irrelevant information, allowing for a more focused retrieval of relevant documents. This strategy can reduce search time and improve document relevance.

\subsubsection{AT-RAG Model}
The \gls{mm} enhances the \glspl{llm} by integrating retrieved documents and performing multi-step reasoning. Formally, let $\rvq$ represent the query and $\mathbf{D}$ denote the external knowledge base. Using a topic assignment model $f_{\boldsymbol{\theta}}(.)$, the associated query topic is defined as $t = f_{\boldsymbol{\theta}}(\rvq)$, where $\boldsymbol{\theta}$ represents the parameters of the model.

The \gls{mm} process begins by passing $\rvq$ through $f_{\boldsymbol{\theta}}(.)$ to generate the topic $t_1$. This topic is then used to retrieve a set of relevant documents $\rvd_1$ from $\mathbf{D}$:
\[
\rvd_1 = \text{Retriever}(\rvq, t_1; \mathbf{D})
\]
Next, the model generates reasoning steps (\gls{cot}), denoted as $\rvc_1$, and integrates these documents into the \gls{llm}'s input to generate an answer $\hat{\rva}$:
\[
\hat{\rva} = \text{LLM}(\rvq, \rvd_1, \rvc_1)
\]
Should the initial retrieval prove inadequate, as determined by the Answer Grader module (refer to section \ref{sec:answergrader} for further information), the system is capable of iteratively enhancing the query and executing the retrieval process again in subsequent iterations:
\[
\hat{\rva}_i = \text{LLM}(\rvq_i, \rvd_i, \rvc_{1:i})
\]
where $\rvq_i$ is the refined query at iteration $i$, and $\rvc_{1:i}$ is the accumulated reasoning context. Documents are retrieved as follows:
\[
\rvd_i = \text{Retriever}(\rvq_i, t_i; \mathbf{D})
\]
where $t_i$ is generated by $t_i = f_{\boldsymbol{\theta}}(\rvq_i)$.
\subsubsection{Topic Assignment Model}
The topic assignment model enhances retrieval accuracy by refining the search space. It predicts the most relevant topic for a given query, enabling the retrieval system to focus on a specific subset of the knowledge base. Let $f_{\boldsymbol{\theta}}(.)$ denote the topic assignment model. For a query at iteration $i$, $\rvq_i$, the corresponding topic $t_i$ is generated as:
\[
t_i = f_{\boldsymbol{\theta}}(\rvq_i)
\]
The topic $t$ summarizes the query's domain, filtering the document database $\mathbf{D}$ to improve retrieval precision and reduce computational complexity.

We implement this using BERTopic \citep{grootendorst2022bertopic}, which leverages transformer-based models for advanced topic discovery. BERTopic clusters document embeddings and applies a class-based model to create coherent topic representations, effectively capturing contextual word meanings.

For each multi-hop dataset, we fine-tune a pre-trained BERTopic model and apply it during both inference and data ingestion phases. This fine-tuning process adapts the model to each dataset's unique characteristics, enhancing topic coherence and retrieval accuracy.

\subsubsection{Analysis of Topic Distribution Across Datasets}
\begin{figure}[tb]
\includegraphics[width=.8\columnwidth]{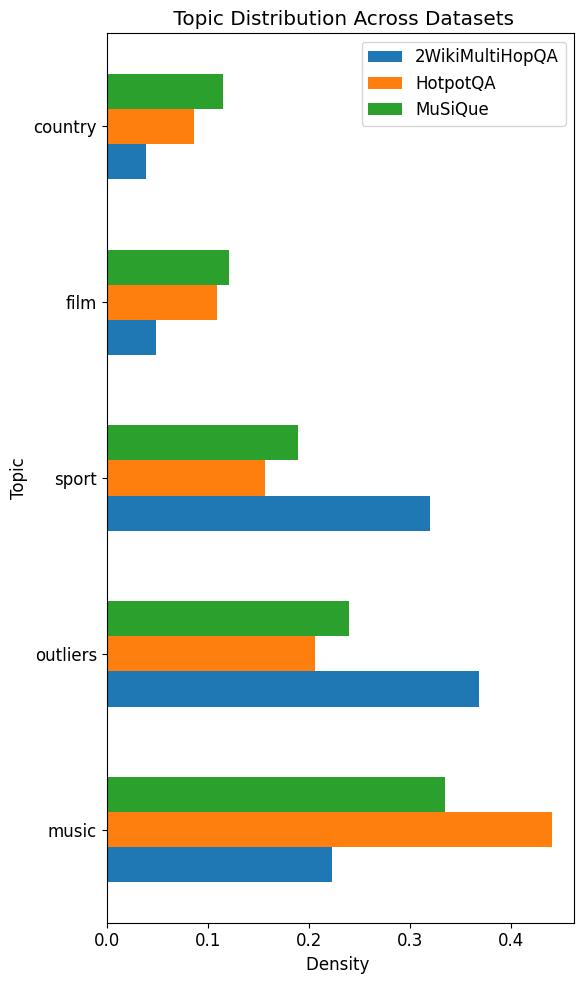}
    \caption{Normalized Topic Distribution Using TopicBERT Across Multi-Hop QA Datasets. The bar plot displays the relative density (proportion of total documents) for each topic, highlighting thematic diversity within each dataset. This visualization emphasizes how topic assignment addresses dataset bias, influencing the retrieval process in \gls{qa} tasks.}
    \label{fig:performance_topic}
\end{figure}

To highlight the importance of topic assignment in \gls{mm}, we analyzed the topic distribution for documents in each multi-hop dataset. We fine-tuned the BertTopic model for the datasets and visualized the distribution of the top 5 topics. Figure \ref{fig:performance_topic} illustrates how topic assignment helps mitigate dataset bias, which can influence retrieval processes and \gls{rag} model performance.

Our analysis reveals distinct differences in topic distribution across datasets. For instance, "film" content is similarly represented in \textbf{MuSiQue} and \textbf{HotpotQA}, but less prevalent in \textbf{2WikiMultiHopQA}. "Music" topics are more prominent in \textbf{MuSiQue}, while \textbf{HotpotQA} and \textbf{2WikiMultiHopQA} show less emphasis on this area.

Understanding these distributions is crucial for identifying dominant themes and thematic focus in each dataset. This insight is vital for improving retrieval performance and allows for tailoring the retriever to match the topic distribution, ensuring more relevant document selection for each query.

\subsubsection{Data Ingestion for Vector Database}

The data ingestion process prepares the dataset for embedding and vector store creation. First, documents and metadata are extracted. Topic assignment is performed using a specialized model, assigning topics and probabilities to each document and adding them to the metadata. Using an embedding model, the vector embedding step converts documents into dense vector embeddings. These embeddings are stored in a vector database (e.g., Chroma) for efficient similarity search. Documents are processed in batches, and the vector store is persisted for future retrieval, enabling content- and topic-based filtering.

\subsubsection{Answer Grader}
\label{sec:answergrader}
As illustrated in Figure \ref{fig:model}, \gls{mm} incorporates an Answer Grader module to evaluate the quality and relevance of generated responses. This module consists of two key components:

The Usefulness Grader: Assesses the relevance and value of the answer to the user's query.
The Hallucination Grader: Verifies the factual accuracy of the answer by cross-referencing it with the retrieved documents, minimizing the risk of hallucinations.

If discrepancies are detected, the RewriteQuery module is activated to reformulate the original query, addressing information gaps or ambiguities. The reformulated query is then passed back to the retriever for additional context.
To prevent endless querying, we implement a maximum iteration limit, N. If reached, the process terminates, outputting the final state answer.
All three modules—Usefulness Grader, Hallucination Grader, and RewriteQuery—utilize prompting techniques with \gls{llm}s for efficient and effective processing.
For a comprehensive understanding of the entire \gls{qa} process using \gls{mm}, refer to Algorithm \ref{alg:atrag}, which provides a detailed step-by-step workflow breakdown.

\begin{algorithm}
\small
\caption{\gls{mm} Model Inference}
\begin{algorithmic}[1]
\Require Query $\rvq$, Knowledge base $\mathbf{D}$, Topic assignment model $f_{\boldsymbol{\theta}}$, Max iterations $N$
\Ensure Final answer $\hat{\rva}$
\State Initialize context $\rvq_1 \gets \rvq$
\For{$i = 1$ to $N$}
    \State Generate topic: $t_i \gets f_{\boldsymbol{\theta}}(\rvq_i)$
    \State Retrieve documents: $\rvd_i \gets \text{Retriever}(\rvq_i, t_i; \mathbf{D})$
    \State Generate reasoning steps: $\rvc_i \gets \text{CoT}(\rvq_i, \rvd_i)$
    \State Produce answer: $\hat{\rva}_i \gets \text{LLM}(\rvq, \rvd_i, \rvc_{1:i})$
    \If{UsefulnessGrader($\hat{\rva}_i$) is satisfactory}
        \If{HallucinationGrader($\hat{\rva}_i, \rvd_i$) is not hallucinating}
            \State \Return $\hat{\rva}_i$
        \EndIf
    \EndIf
    \State Update query: $\rvq_{i+1} \gets \text{RewriteQuery}(\rvq_i, \hat{\rva}_i, \rvc_i)$
\EndFor
\State \Return $\hat{\rva}_N$
\end{algorithmic}
\label{alg:atrag}
\end{algorithm}

%% file: sec-experiments.tex
\section{Experiments}
To benchmark the \gls{mm} model and rigorously evaluate its effectiveness in handling complex queries, we test it on multi-hop \gls{qa} datasets. To address more challenging query scenarios, we employ three benchmark multi-hop \gls{qa} datasets that require sequential reasoning across multiple documents: 1) MuSiQue \citep{trivedi2022interleaving}, 2) HotpotQA \citep{yang2018enhancing}, and 3) 2WikiMultiHopQA \citep{ho2020constructing}.

\subsection{Multi-hop QA Dataset}
The evolution of question-answering systems has led to the creation of multi-hop datasets designed to challenge and evaluate more advanced \gls{qa} models. These datasets, including MuSiQue, HotpotQA, and 2WikiMultiHopQA, feature complex queries that require reasoning across multiple documents. Unlike single-hop \gls{qa}, where answers can be extracted from a single source, multi-hop \gls{qa} necessitates synthesizing information from various sources to generate accurate answers.

MuSiQue \citep{trivedi2022interleaving} focuses on questions requiring multistep reasoning by combining multiple facts. This dataset is particularly valuable for assessing models' ability to navigate interconnected information and draw logical conclusions. HotpotQA \citep{yang2018enhancing} emphasizes both reasoning and fact verification by querying linked documents. It tests a model's ability to find the correct answer and provide supporting facts, thereby assessing comprehension and inference skills simultaneously.

2WikiMultiHopQA \citep{ho2020constructing} leverages linked Wikipedia articles to challenge models with reasoning paths that span multiple documents. This dataset is useful for evaluating how well \gls{qa} systems handle real-world knowledge structures and navigate interlinked information sources.

The primary difference between these multi-hop datasets and their single-hop counterparts is the requirement for information synthesis. Multi-hop questions cannot be answered without combining and reasoning over information from multiple documents or data points. This characteristic makes these datasets essential for evaluating advanced, inference-driven models that emulate human-like reasoning processes.

As \gls{qa} models continue to advance, these multi-hop datasets play a crucial role in pushing the boundaries of machine comprehension and reasoning, driving the development of more sophisticated \gls{qa} systems capable of handling real-world complexity.

\subsection{LLMs as Autonomous Judges in QA Evaluation}
\begin{table*}[t!]
\caption{Comparison between different RAG approaches.}
\vspace{-0.1in}
\label{tab:main}

\centering
\resizebox{\textwidth}{!}{
\renewcommand{\arraystretch}{1.0}
\begin{tabular}{lcccccccccccc}
\toprule

& \multicolumn{4}{c}{\bf 2WikiMultiHopQA} & \multicolumn{4}{c}{\bf HotpotQA} & \multicolumn{4}{c}{\bf MuSiQue} \\
\cmidrule(l{2pt}r{2pt}){2-5} \cmidrule(l{2pt}r{2pt}){6-9} \cmidrule(l{2pt}r{2pt}){10-13}

\textbf{Approach} & Corr. & Comp. & Rel. & Overall & Corr. & Comp. & Rel. & Overall & Corr. & Comp. & Rel. & Overall \\
\midrule

\textbf{One Step RAG} 
& 3.84 & 3.94 & 7.03 & 4.94 & 5.93 & 5.82 & 7.97 & 6.57 & 3.00 & 3.12 & 5.38 & 3.83 \\
\noalign{\vskip 0.25ex}\cdashline{1-13}\noalign{\vskip 0.75ex}

\textbf{Adaptive RAG} 
& \bf5.99 & 5.00 & \bf8.18 & 6.39 & 6.01 & 5.55 & 7.72 & 6.43 & 3.64 & 3.63 & 5.88 & 4.38 \\
\noalign{\vskip 0.25ex}\cdashline{1-13}\noalign{\vskip 0.75ex}

\textbf{\gls{mm} (with GPT-4o) } 
& 5.79 & \bf5.72 & \bf8.18 & \bf6.57 & \bf7.27 & \bf6.98 & \bf8.56 & \bf7.61 & \bf3.65 & \bf3.88 & \bf6.02 & \bf4.52 \\

\bottomrule

\end{tabular}
}
\end{table*}
The integration of \glspl{llm} as autonomous judges in \gls{qa} evaluation has transformed the automation of assessing responses based on qualitative metrics. By leveraging the deep comprehension and reasoning capabilities of models like GPT-4o, \gls{llm}-based evaluation systems can provide detailed insights into the correctness, completeness, relevance, and clarity of generated answers.

To automatically evaluate \gls{qa} pairs (both generated and ground truth), an \gls{llm} such as GPT-4o is used as the judge. The process begins by presenting the \gls{llm} with a question, the ground truth answer, and the generated response\citep{badshah2024reference}. The model is prompted to assess the generated answer based on the following predefined criteria:

\begin{enumerate}
    \item \textbf{Correctness}: Does the generated answer accurately align with the ground truth?
    \item \textbf{Completeness}: Does the response include all necessary information relevant to the question?
    \item \textbf{Relevance}: Is the answer relevant to the posed question?
\end{enumerate}

For each criterion, the \gls{llm} assigns a score between 0 and 10, with the overall score calculated as the average of the individual scores. This approach enables a nuanced evaluation, going beyond token-level comparison and allowing for a more human-like understanding of the content.

\subsection{Experimental Results and Analyses}
In this section, we compare the performance of three different \gls{rag} approaches: One Step \gls{rag}, Adaptive \gls{rag}, and our proposed \gls{mm} method. The evaluation criteria include correctness, completeness, relevance, and overall score, as shown in Table \ref{tab:main}.

As independent evaluators, the experimental results were analyzed using \glspl{llm} on a subset of 500 \gls{qa} samples from each dataset. The answers' correctness, completeness, and relevance were scored using GPT-4, each score ranging from 0 to 10, and the overall score was calculated as the average of these metrics. This approach goes beyond traditional token-level evaluation, leveraging the deep reasoning capabilities of the \gls{llm} to simulate human-like judgment.

As demonstrated in Table \ref{tab:main}, the \gls{mm} method consistently outperforms the other two approaches across all datasets. For example, in the 2WikiMultiHopQA dataset, the \gls{mm} method achieved a correctness score of 5.79, a completeness score of 5.72, and a relevance score of 8.18, resulting in an overall score of 6.57. This is a significant improvement over both One Step \gls{rag} and Adaptive \gls{rag}, with the latter scoring 6.39 overall.

A similar trend is observed in the HotpotQA dataset, where the \gls{mm} method achieved an overall score of 7.61, compared to 6.57 for One Step \gls{rag} and 6.43 for Adaptive \gls{rag}. The MuSiQue dataset further highlights the efficacy of the \gls{mm} approach, with an overall score of 4.52, outperforming both One Step and Adaptive \gls{rag} methods.

These results show that the \gls{mm} method, which incorporates topic filtering, improves the quality of the answers by focusing on topic-relevant documents. This leads to more accurate, complete, and relevant responses, ultimately improving the overall performance of the \gls{qa} system.

\subsubsection{Ablation Study} 
To further assess the robustness of our proposed \gls{mm} method, we conducted an ablation study using different \gls{llm}s, including GPT-4o and Mixtral8x7B \citep{jiang2024mixtral}. This study aims to understand how the performance of the \gls{mm} approach varies between different \gls{llm}s in multi-hop \gls{qa} datasets. The results are summarized in Table \ref{tab:ablation}.
In all three datasets, GPT-4o consistently outperformed Mixtral8x7B across all evaluation metrics, demonstrating superior performance overall. Although both models performed lower in the MuSiQue dataset, GPT-4o still maintained an edge over Mixtral8x7B \citep{jiang2024mixtral}.
The ablation study clearly demonstrates that the choice of \gls{llm} significantly impacts the performance of the \gls{mm} method.

\begin{table*}[t!]
\caption{Comparison between different \gls{llm}s with \gls{mm} model.}
\vspace{-0.1in}
\label{tab:ablation}

\centering
\resizebox{\textwidth}{!}{
\renewcommand{\arraystretch}{1.0}
\begin{tabular}{lcccccccccccc}
\toprule

& \multicolumn{4}{c}{\bf 2WikiMultiHopQA} & \multicolumn{4}{c}{\bf HotpotQA} & \multicolumn{4}{c}{\bf MuSiQue} \\
\cmidrule(l{2pt}r{2pt}){2-5} \cmidrule(l{2pt}r{2pt}){6-9} \cmidrule(l{2pt}r{2pt}){10-13}

\textbf{Approach} & Corr. & Comp. & Rel. & Overall & Corr. & Comp. & Rel. & Overall & Corr. & Comp. & Rel. & Overall \\
\midrule

\textbf{\gls{mm} (with GPT-4o) } 
& \bf5.79 & \bf5.72 & \bf8.18 & \bf6.57 & \bf7.27 & \bf6.98 & \bf8.56 & \bf7.61 & \bf3.65 & \bf3.88 & \bf6.02 & \bf4.52 \\
\noalign{\vskip 0.25ex}\cdashline{1-13}\noalign{\vskip 0.75ex}
\textbf{\gls{mm} (with Mixtral8x7B) } 
& 3.41 & 3.73 & 6.01 & 4.38 & 6.20 & 6.02 & 7.57 & 6.60 & 2.81 & 3.11 & 5.34 & 3.75 \\

\bottomrule
\end{tabular}
}
\end{table*}

%% file: sec-application.tex
\section{Application on Medical QA}

We conducted a case study to assess the effectiveness of our proposed method in answering multi-hop, time-based queries from medical records. The medical records of the patient, which include a variety of longitudinal documents such as doctor's notes, lab results, and diagnostics, pose a challenge for the \gls{rag} frameworks. For instance, the query: \textit{What are the abnormal lab results of the patient within the last year?} is difficult for One Step \gls{rag} approaches, which must retrieve relevant chunks from vast data and may lack reasoning to select the correct time-stamped information. 

\begin{figure}[tb]

\includegraphics[width=.95\columnwidth]{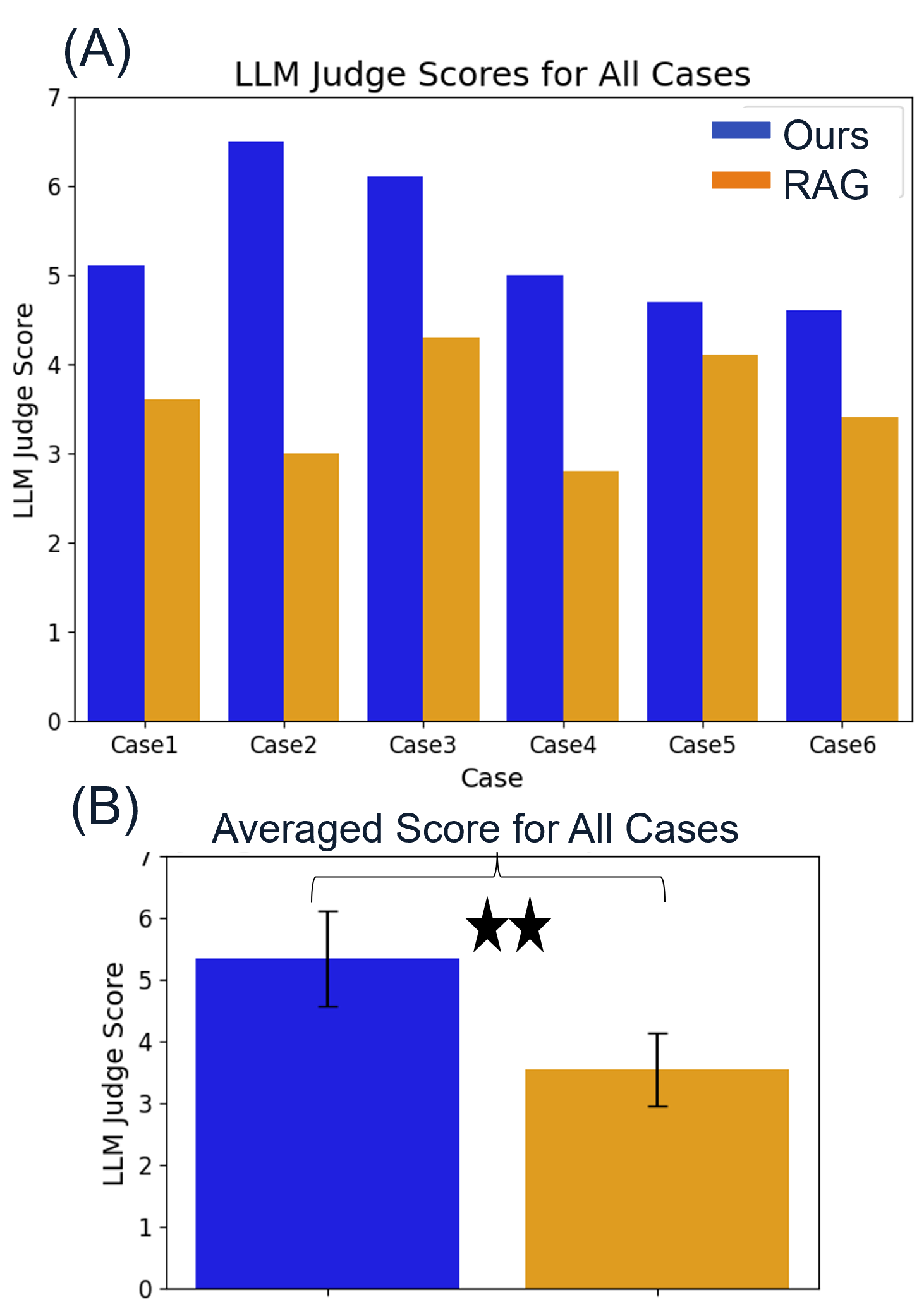}
    \caption{\gls{mm} (A) A comparison between our \gls{mm} and the One Step \gls{rag} in answering time-based questions on the medical records of six patients evaluated by GPT-4. (B) The average scores of our proposed approach and the One Step \gls{rag} across the six cases. ** indicates a statistically significant difference between the two bars, with p < 0.02 as determined by ANOVA test \citep{st1989analysis}. 
    }
    \label{fig:medical}
\end{figure}

We assessed our \gls{mm} method by comparing it to the One Step \gls{rag} within a dataset of medical records from six patients. As shown in Figure \ref{fig:medical}, \gls{mm} outperformed One Step \gls{rag} in all cases. Using Mixtral8x7B \citep{jiang2024mixtral} as the \gls{llm} and GPT-4o as the \gls{llm} judge, we queried each case with ten time-based questions. Our method attained an average score of 5.3, nearly two points above the One Step \gls{rag}, which had an average score of 3.5.

Table \ref{tab:qa1} illustrates a query on abnormal HbA1c values over the past two years. Our fine-tuned topic assignment model for this dataset identified the topic as LabResult, narrowing the retrieval to lab data. A \gls{cot} prompt further refined the search by focusing on the correct time range and the HbA1c condition, resulting in a correct answer, while naive \gls{rag} provided an incorrect one.

Table \ref{tab:qa2} shows another question about the last doctor visit and its reason. The topic assignment model assigned ClinicalNote to the query, limiting the search to relevant clinical notes. The \gls{cot} prompt guided the \gls{llm} to identify the date and reason for the visit. Our method retrieved the correct details, while naive \gls{rag} mistakenly returned a lab test date with no valid reason for the visit.

%% file: sec-discussion.tex
\section{Conclusion}
In this paper, we proposed \gls{mm}, a novel \gls{rag} model designed to tackle complex multi-hop \gls{qa} tasks. By integrating topic assignment through BERTopic, we significantly improved the speed and accuracy of multi-step document retrieval. The model effectively combines topic filtering with \gls{cot} reasoning to reduce the search space and focus retrieval on the most relevant documents. Experimental results on multi-hop \gls{qa} demonstrated that \gls{mm} outperforms existing \gls{rag} approaches in terms of correctness, completeness, relevance, and time efficiency. These improvements suggest that our approach is a promising solution for handling intricate \gls{qa} tasks.

\section{Limitations}
Despite promising results, \gls{mm} has some limitations that warrant further investigation. First, the performance of the model is highly dependent on the quality of the initial topic assignment. If the assigned topic is incorrect or too broad, the retrieval precision may decrease, leading to less accurate answers. Second, while we demonstrated improvements in retrieval efficiency, the multi-step nature of our approach can still be computationally expensive, particularly when applied to large datasets or real-time systems. Future work could explore methods to dynamically adjust the topic assignment model based on query complexity or integrate adaptive topic assigners mechanisms to further optimize the search process.

\section{Ethics Statement}
The development and deployment of \gls{llm}s such as those used in \gls{mm} must be guided by strong ethical considerations. While our model aims to improve information retrieval and \gls{qa}, we recognize the potential risks associated with \gls{llm}s, including the possibility of generating incorrect or biased information. In particular, the use of external knowledge sources raises concerns about the credibility and accuracy of retrieved documents, especially in critical domains such as healthcare as investigated here. It is crucial to ensure that the information retrieved by the model is reliable and factually grounded, and that any potential biases in the underlying datasets are mitigated.

%% file: sec-appendix.tex
\section{Appendix}
\subsection{Sample Quality Examples for Medical QA}
\begin{table*}[ht]
\begin{tabular}{|>{\raggedright}p{5cm}|>{\raggedright}p{5cm}|>{\raggedright\arraybackslash}p{5cm}|}
\hline
\textbf{Aspect} & \textbf{Simple Q\&A} & \textbf{Q\&A Graph} \\
\hline
\multirow{2}{5cm}{\textbf{query}} 
    & \multicolumn{2}{>{\raggedright\arraybackslash}p{10cm}|}{Does the patient have a history of HbA1c>6.0\% within the past 2 years?} \\
    & &\\
\hline
% \multirow{2}{5cm}{\textbf{ground-truth}} 
%     & \multicolumn{2}{>{\raggedright\arraybackslash}p{10cm}|}{No, the patient does not have any history of HbA1c above 6\%} \\
%     & &\\

% \textbf{generated answer} 
%     & Yes, HbA1c>6.0 within past 2 years 
%     & Based on the provided context, most recent HbA1c value was 5.2\%, which is below the threshold of 6.0 you asked about. This result is from July 25, 2022. \\
\hline
\textbf{LLM Judge Score} 
    & 0 
    & 10 \\
\hline
\textbf{Assigned Topic} 
    & - 
    & LabResult \\
\hline
\textbf{Chain of Thought (CoT)} 
    & - 
    & To determine if the patient has a history of HbA1c>6.0 within the past 2 years, we need to search for any lab results related to HbA1c in October 2022 or after. If there is a result with HbA1c>6.0, then the patient has a history of HbA1c>6.0 within the past 2 years. \\
\hline
\end{tabular}
\caption{A comparison was made between One Step \gls{rag} and \gls{mm} concerning the most recent doctor visit. As the \gls{qa} process was conducted on medical records, we omitted the final answer and focused solely on the \gls{llm} Score to evaluate \gls{mm} relative to One Step \gls{rag}.}
\label{tab:qa1}
\end{table*}

\begin{table*}[ht]
\begin{tabular}{|>{\raggedright}p{5cm}|>{\raggedright}p{5cm}|>{\raggedright\arraybackslash}p{5cm}|}
\hline
\textbf{Aspect} & \textbf{Simple Q\&A} & \textbf{Q\&A Graph} \\
\hline
\multirow{2}{5cm}{\textbf{query}} 
    & \multicolumn{2}{>{\raggedright\arraybackslash}p{10cm}|}{When was the last doctor visit and the reason behind it?} \\
    & & \\
\hline
% \multirow{2}{5cm}{\textbf{ground-truth}} 
%     & \multicolumn{2}{>{\raggedright\arraybackslash}p{10cm}|}{12/12/2022, follow up for diabetes} \\
%     & & \\
% \hline
% \textbf{generated answer} 
%     & The last doctor visit was on 01/2023 for a routine visit. 
%     & The last doctor visit was on 12/12/2022 and it was for a follow-up related to diabetes. \\

\textbf{LLM Judge Score} 
    & 0 
    & 10 \\
\hline
\textbf{Assigned Topic} 
    & - 
    & ClinicalNote \\
\hline
\textbf{Chain of Thought (CoT)} 
    & - 
    & To answer the question, we need to find the most recent clinical note that mentions a doctor visit. We will then extract the date of the visit and the reason behind it from the note. \\
\hline
\end{tabular}
\caption{A comparison was made between One Step \gls{rag} and \gls{mm} concerning the most recent doctor visit. As the \gls{qa} process was conducted on medical records, we omitted the final answer and focused solely on the \gls{llm} Score to evaluate \gls{mm} relative to One Step \gls{rag}.}
\label{tab:qa2}
\end{table*}